# Taming the ReLU with Parallel Dither in a Deep Neural Network

Andrew J.R. Simpson [#1]

[#] *Centre for Vision, Speech and Signal Processing, University of Surrey*
*Surrey, UK*
[1] `Andrew.Simpson@surrey.ac.uk`

*Abstract*—*Rectified Linear Units* (ReLU) seem to have displaced traditional 'smooth' nonlinearities as *activation-function-du-jour* in many – but not all - deep neural network (DNN) applications. However, nobody seems to know *why*. In this article, we argue that ReLU are useful because they are ideal demodulators – this helps them perform fast abstract learning. However, this fast learning comes at the expense of serious nonlinear distortion products - *decoy features*. We show that *Parallel Dither* acts to suppress the *decoy features*, preventing overfitting and leaving the *true features* cleanly demodulated for rapid, reliable learning.

*Index terms*—**Deep learning, ReLU, regularisation, dropout, dither, parallel dither.**

## I. INTRODUCTION

Deep neural networks are able to learn abstract, hierarchical representations through a process of demodulation [1]. For symmetrical carrier signals, demodulation is the result of rectification [1]. In this context, it is not surprising that the *Rectified Linear Unit* (ReLU) [2] has found success in DNNs [3] – it is, in principle, a nearly ideal demodulator function. However, in practice, ReLU have proved somewhat capricious and can even fail outright in some applications. Very little is known about why.

The discrete signal processing interpretation of DNN [1,4] views each neuron as a step of abstraction. The weights of a neuron are viewed as linear filter coefficients and the subsequent activation function as a demodulating nonlinearity. In this context, the goal of training is to learn the hierarchical set of filters whose demodulated outputs best capture the abstract structure of the training and test data. In this context, through gradient descent, the DNN is looking for those filters [4] which most easily allow discrimination. As a result, one cause of overfitting in a DNN would be the learning of filters which capture *decoy features* (artefacts of the data) rather than capturing the true (i.e., general) features of what is represented in the data.

Following this line of reasoning, even for artefact-free data, nonlinear distortion introduced by the activation function might provide *decoy features* [4]. For example, if the true features of the data exist in some particular region of the feature space and the nonlinear activation function projects distortion products (harmonic or intermodulation) into a less dense region of the feature space, then filters may more easily be learned which exploit these *decoy features*.

The learning of filters capturing decoy features is potentially a problem for a number of reasons. Firstly, artefacts of any kind are unlikely to generalise (e.g., to the test data). Secondly, the decoy features may not necessarily be reliably demodulated further up the hierarchy, especially if they are either high-order or the result of aliasing [4]. As the ReLU function [$max(0, x)$] is extremely abrupt, it must introduce such decoy features and hence cause overfitting.

In signal processing, dither is used to suppress (decorrelate) nonlinear distortion products. Dither also acts to regularise DNN by the same means [6,7]. In this article, we show that dither also works to regularise ReLU DNNs and we illustrate the corresponding effect on demodulation.

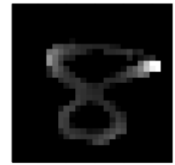

Fig. 1. **Example MNIST image.** We took the 28x28 pixel images and unpacked them into a vector of length 784 to form the input at the first layer of the DNN.

## II. METHOD

*Demodulation via ReLU.* In order to illustrate the demodulation facility of ReLU we consider an intuitive example case in the audio domain. We constructed a simple 10-second test signal, sampled at 44100 cycles/s, featuring a sinusoidal carrier that was modulated by (multiplied with) a sinusoidal modulator. The carrier was at a frequency of 10000 cycles/s and the modulator was at a frequency of 100 cycles/s. Multiplied together, this resulted in an amplitude modulated carrier. The task for a demodulator is to extract, from the carrier, the modulation signal. Fig. 2a plots the power spectral density for the test signal passed through the ReLU. To apply parallel dither, the test signal was replicated 100x. Uniform random noise (zero mean, unit scale) was added to each independent instance of the test signal and the result was passed through the ReLU. The resulting 100 processed test signals were then averaged. Fig. 2b plots the resulting power spectral density for the parallel dithered ReLU. In both plots, the strong demodulated energy at 100 cycles illustrates the capacity of the ReLU to demodulate. However, in Fig. 2a, the other large peaks in the spectrum are distortion and, hence, represent potential *decoy features*. In Fig. 2b, the parallel-dithered power spectrum is slightly noisier but the suppression of distortion products is clear – there are no *decoy features*. Thus, the argument is simple: without *decoy features*, learning cannot be led astray and filters must capture true features, thus reducing the problem of overfitting.

*DNN with ReLU*. In order to illustrate how the process of dithering improves the cause for ReLU in the DNN context, we use the well-known computer vision problem of hand-written digit classification using the MNIST dataset [7]. For the input layer we unpacked the images of 28x28 pixels into vectors of length 784. An example digit is given in Fig. 1. Pixel intensities were normalized to zero mean. Replicating Hinton's [8] architecture, but using ReLU, we built a fully connected network of size 784x100x10 units, where the 10-unit softmax output layer corresponds to the 10-way digit classification problem.

Operating within the so-called 'small-data regime' (as in [5,6]), we used only the first 256 training examples of the MNIST dataset and tested on the full 10,000 test examples. We trained several instances of the model with non-batch SGD (equivalent to a batch size of 1 in batch-averaged SGD). The first was a baseline model without regularisation. The second was the baseline model regularised with dropout. The third was the baseline model regularised with 100x parallel dither [6]. The fourth was the baseline model regularised using 100x parallel dither w/ dropout [6,9].

*Parallel dither and dropout*. During non-batch SGD, each training example was replicated 100 times to form a parallel set. For parallel dither, each element of this set was dithered independently by adding uniform random noise of zero mean and unit scale and the gradients computed for each element independently. For parallel dither w/ dropout, both dither and dropout were applied at the same time (i.e., the parallel set was still of size: 100). Then, each parallel set of gradients (representing a single training example) was averaged and applied. Batch averaging across training examples was not applied.

Each separate instance of the model was trained for 100 full-sweep iterations of non-batch SGD (without momentum) and the test error computed (over the 10,000 test examples) at each iteration. For reliable comparison, each instance of the model was trained from the exact same random starting weights. A learning rate (SGD step size) of 0.01 (as was optimal for ReLU) was used for all training. All dropout was at the 50% level. These parameter choices allow useful comparison with the equivalent optimally-biased-sigmoid [1] results of [6] (which were trained from the exact same random starting weights on the same problem). The biased-sigmoid data of [6], for the model regularised with 100x parallel dither w/ dropout, is included both for reference and because it is explicitly optimised for demodulation [1].

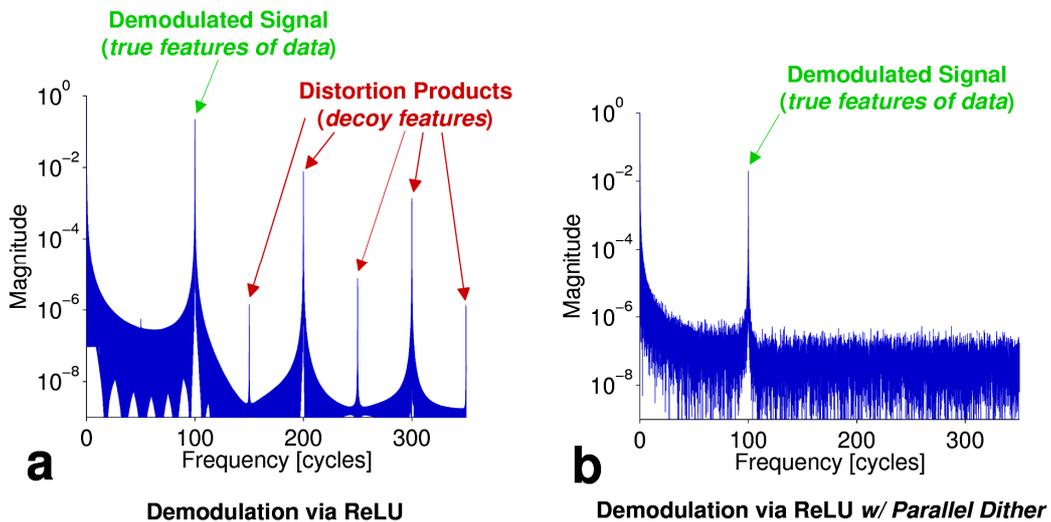

**Fig. 2.** *True features and decoy features*: demodulation via ReLU with and without *Parallel Dither*. A audio-domain test signal was constructed with a sinusoidal carrier signal (10000 cycles/s) modulated by (multiplied with) a 100 cycle/s sinusoidal modulator (similar to [1]). The demodulation (rectification) process extracts the envelope modulations applied to the carrier. **a** plots the power spectral density for the test signal demodulated with ReLU and **b** plots the same for parallel-dithered [see 6] ReLU.

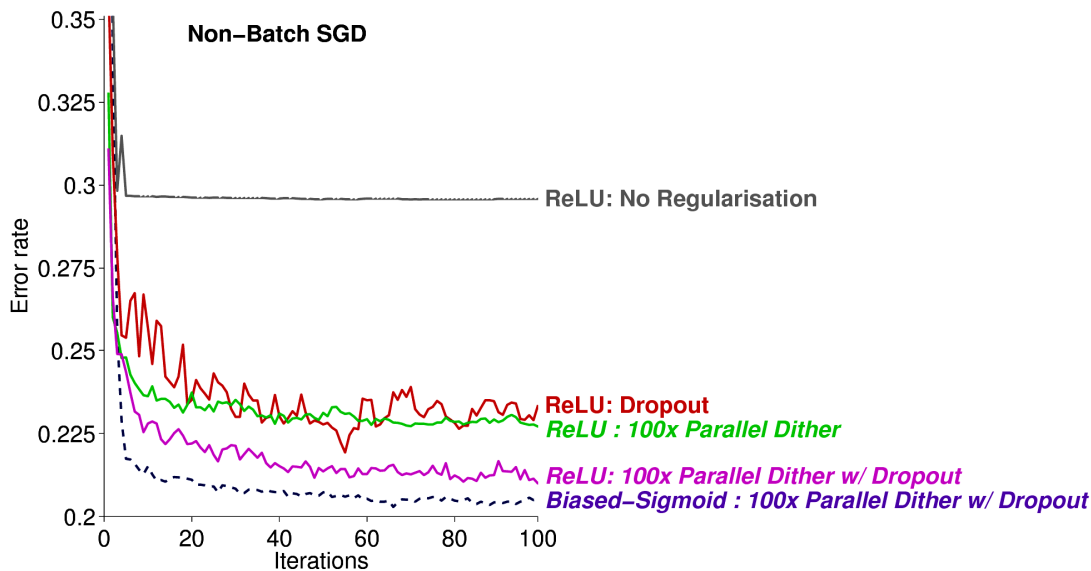

**Fig. 3. ReLU versus Parallel Dither.** Test error functions of (non-batch) SGD iterations, for the various ReLU models, including the equivalent biased-sigmoid data of [6] for comparison. All models were trained on the same data and from the same random starting weights. *NB: The y-axis is somewhat cropped for better scale.*

III. RESULTS

Fig. 3 plots the test-error rates, as a function of full-sweep SGD iterations, for the various non-batch-SGD trained models. The ReLU model trained without regularisation performs poorly. The model trained with dropout fares better and the model trained with 100x parallel dither converges similarly but learns faster (than dropout). The model trained with parallel dither w/ dropout performs best (as in [6]).

We also include plotted here (Fig. 3) equivalent data (from [6]) from a model with the same architecture but featuring the biased-sigmoid [1] activation function, trained on the same data and from the same random starting weights. The same 100x parallel dither w/ dropout was also applied in [6]. This provides an interesting comparison because the biased sigmoid [1] activation function of this model was explicitly optimised for demodulation but is much smoother than the ReLU. Clearly, the optimally-biased-sigmoid DNN learns faster and performs better than the equivalent ReLU DNN.

We also note that the ReLU models trained best at a learning rate of 0.01 (as plotted here), whereas the identical biased-sigmoid model trained faster at a learning rate of 1 (which was much too high to train at all with any of the ReLU models). Taken together, this tends to suggest that the raw-demodulation-power advantage of ReLU (over traditional smooth activation functions) does not hold up if the smooth activation function is biased for optimum demodulation [e.g., 1]. Hence, the remaining advantage of ReLU appears to be cheap computation. Furthermore, given the large magnitude *decoy features* produced by the ReLU, it is not surprising that ReLU have not found favour in recurrent neural networks (though parallel dither [6] might help as in [10,11]).

IV. DISCUSSION AND CONCLUSION

In this paper, following the discrete signal processing interpretation of deep neural networks [1,4-6], we have interpreted ReLU in terms of demodulation and we have introduced the concept of *decoy features* to capture the action of a DNN learning artefactual features produced by abrupt activation-function nonlinearities. We have illustrated the ability of parallel dither [6,5] to suppress these decoy features and hence to regularise a ReLU DNN. We have also shown that the demodulation advantge of the ReLU over the traditional smooth activation functions is reversed when compared with the biased-sigmoid [1] (which is explicitly optimised for demodulation).

ACKNOWLEDGMENT

AJRS did this work on the weekends and was supported by his wife and children.